\documentclass[runningheads]{llncs}
\usepackage{graphicx}
\usepackage{comment}
\usepackage{amsmath,amssymb} 
\usepackage{color}

\usepackage{tablefootnote}
\usepackage{subcaption}
\usepackage{graphicx}
\usepackage{verbatim}
\usepackage{color}
\usepackage{booktabs}
\graphicspath{{images/}}
\usepackage{float}
\usepackage{enumitem}

\usepackage[pagebackref=true,breaklinks=true,letterpaper=true,colorlinks,bookmarks=false]{hyperref}

\newcommand{\figref}[1]{Fig.~\ref{#1}}
\newcommand{\secref}[1]{Sec.~\ref{#1}}
\newcommand{\tabref}[1]{Table~\ref{#1}}
\newcommand{\with}[2]{{#1} + {#2}}
\newcommand{\without}[1]{-- \emph{without} {#1}}

\newcommand{\eg}{\textit{e.g.}\ }
\newcommand{\ie}{\textit{i.e.}\ }

\begin{document}
\pagestyle{headings}
\mainmatter
\def\ECCVSubNumber{5457}  

\title{SoftPoolNet: Shape Descriptor for Point Cloud Completion and Classification} 

\titlerunning{SoftPoolNet}
%
\author{Yida Wang\inst{1} \and
David Joseph Tan\inst{2} \and
Nassir Navab\inst{1} \and
Federico Tombari\inst{1,2}}
\authorrunning{Y. Wang et al.}
%
\institute{Technische Universit\"at M\"unchen\and
Google Inc.}

\maketitle

\begin{abstract}
	Point clouds are often the default choice for many applications as they exhibit more flexibility and efficiency than volumetric data. 
	Nevertheless, their unorganized nature -- points are stored in an unordered way -- makes them less suited to be processed by deep learning pipelines.
	In this paper, we propose a method for 3D object completion and classification based on point clouds.
	We 
	introduce a new way of organizing the extracted features based on their activations, which we name soft pooling.
	For the decoder stage, we propose regional convolutions, a novel operator aimed at maximizing the global activation entropy.
	Furthermore, inspired by the local refining procedure in Point Completion Network (PCN), we also propose a patch-deforming operation to simulate deconvolutional operations for point clouds.
	This paper proves that our regional activation can be incorporated in many point cloud architectures like AtlasNet and PCN, leading to better performance for geometric completion.
	We evaluate our approach on different 3D tasks such as object completion and classification, achieving state-of-the-art accuracy.
\end{abstract}

\section{Introduction}

%
Point clouds are unorganized sparse representations of a 3D point set. Compared to other common representations for 3D data such as 3D meshes and voxel maps, they are simple and flexible, while being able to store fine details of a surface. For this reason, they are frequently employed for many applications within 3D perception and 3D computer vision such as robotic manipulation and navigation, scene understanding, and augmented/virtual reality. Recently, deep learning approaches have been proposed to learn from point clouds for 3D perception tasks such as point cloud classification~\cite{arief2019density,liu2019relation,qi2017pointnet,qi2017pointnet++} or point cloud segmentation~\cite{landrieu2018large,li2018pointcnn,liu2019relation,pham2019jsis3d,wu2019pointconv}. Among them, one of the key breakthroughs in handling unorganized point clouds was proposed by PointNet \cite{qi2017pointnet}, introducing the idea of a max pooling in the feature space to yield permutation invariance. 

\begin{figure}[!t]
\centering
\includegraphics[width=1.0\linewidth]{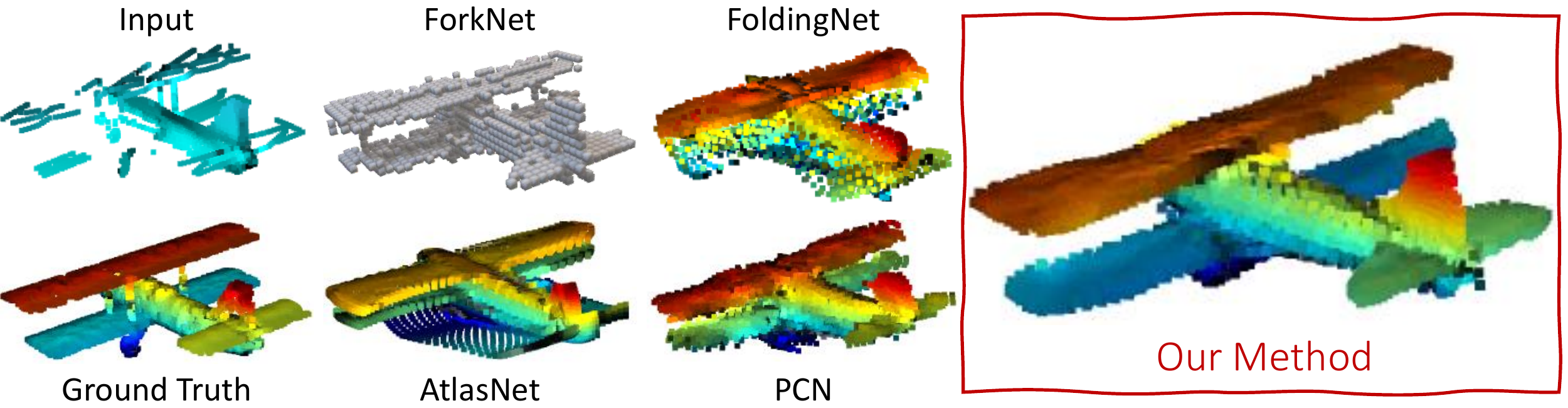}
\caption{This paper proposes a method that reconstructs 3D point cloud models with more fine details. }
\label{fig:teaser}
\end{figure}

An interesting emerging research trend focusing on 3D data is the so-called 3D completion, where the geometry of a partial scene or object acquired from a single viewpoint, \eg through a depth map, is completed of the missing part due to (self-)occlusion as visualized in \figref{fig:teaser}. This can be of great use to aid standard 3D perception tasks such as object modeling, scene registration, part-based segmentation and object pose estimation. Most approaches targeting 3D completion have been proposed for volumetric approaches, since 3D convolutions are naturally suited to this 3D representation. Nevertheless, such approaches bring in the limitations of this representation, including loss of fine details due to discretization and limitations in scaling with the 3D size. Recently, a few approaches have explored the possibility of learning to complete a point cloud~\cite{Groueix_2018_CVPR,yang2018foldingnet,yuan2018pcn}.

This paper proposes an encoder-decoder architecture called SoftPoolNet, which can be employed for any task that processes a point cloud as input in order to regress another point cloud as output. One of the tasks and a main focus for this work is 3D object completion from a partial point cloud.

The theoretical contribution of SoftPoolNet is twofold.
We first introduce soft pooling, a new module that replaces the max-pooling operator in PointNet by taking into account multiple high-scoring features rather than just one. 
The intuition is that, by keeping multiple features with high activations rather than just the highest, we can retain more information while keeping the required permutation invariance.
%
%
A second contribution is the definition of a regional convolution operator that is used within the proposed decoder architecture. This operator is designed specifically for point cloud completion and relies on convolving local features to improve the completion task with fine details.

In addition to evaluating SoftPoolNet for point cloud completion, we also evaluate on the point cloud classification to demonstrate its applicability to general point cloud processing tasks. In both evaluations, SoftPoolNet obtains state of the art results on the standard benchmarks.

\section{Related work}

\paragraph{Volumetric completion.}
Object~\cite{dai2017shape} and scene completion~\cite{song2017semantic,wang2019forknet} are typically carried out by placing all observed elements into a 3D grid with fixed resolution. 
3D-EPN~\cite{dai2017shape} completes a single object using 3D convolutions while 3D-RecGAN~\cite{yang2018dense} further improves the completion performance by using discriminative training. As scene completion contains objects in different scales and more random relative position among all of them, SSCNet~\cite{song2017semantic} proposes a 3D volumetric semantic completion architecture using dilated convolutions to recognize objects with different scales. ForkNet~\cite{wang2019forknet} designs a multi-branch architecture to generate realistic training data to supplement the training.

\paragraph{Point cloud completion.}
Object completion based on point cloud data change partial geometries without using a 3D fixed grid. They represent completed shapes as a set of points with 3D coordinates. For instance, FoldingNet~\cite{yang2018foldingnet} deforms a 2D grid from a global feature such as PointNet~\cite{qi2017pointnet} feature to an output with a desirable shape. AtlasNet~\cite{Groueix_2018_CVPR} generates an object with a set of local patches to simulate mesh data. But  overlaps between different local patches makes the reconstruction noisy. MAP-VAE~\cite{Han_2019_ICCV} predicts the completed shape by joining the observed part with the estimated counterpart.

\paragraph{CNNs for point clouds.}
Existing works like PointConv~\cite{wu2019pointconv} and PointCNN~\cite{li2018pointcnn} index each point with $k$-nearest neighbour search to find local patches, where they then apply the convolution kernels on those local patches. 
Regarding point cloud deconvolutional operations, FoldingNet~\cite{yang2018foldingnet} uses a 2D grid to help generate a 3D point cloud from a single feature. PCN~\cite{yuan2018pcn} further uses local FoldingNet to obtain a fine-grained output from a coarse point cloud with low resolution which could be regarded as an alternative to point cloud deconvolution.


\section{Soft pooling for point features}
\label{sec:soft_pooling}

Given the partial scan of an object, the input to our network is a point cloud with $N_\text{in}$ points written in the matrix form as $\mathbf{P}_\text{in} = [\mathbf{x}_i]_{i = 1}^{N_\text{in}}$ where each point is represented as the 3D coordinates $\mathbf{x}_i = [x_i, y_i, z_i]$.
We then convert each point into a feature vector $\mathbf{f}_i$ with $N_f$ elements by projecting every point with a point-wise multi-layer perceptron~\cite{qi2017pointnet} (MLP) $\mathbf{W}_\text{point}$ with three layers.
Thus, similar to $\mathbf{P}_\text{in}$, we define the $N_\text{in} \times N_f$ feature matrix as $\mathbf{F} = [\mathbf{f}_i]_{i = 1}^{N_\text{in}}$. Note that we applied a softmax function to the output neuron of MLP so that the elements in $\mathbf{f}_i$ ranges between 0 and 1.

\begin{figure}[!t]
\centering
\includegraphics[width=1.0\linewidth]{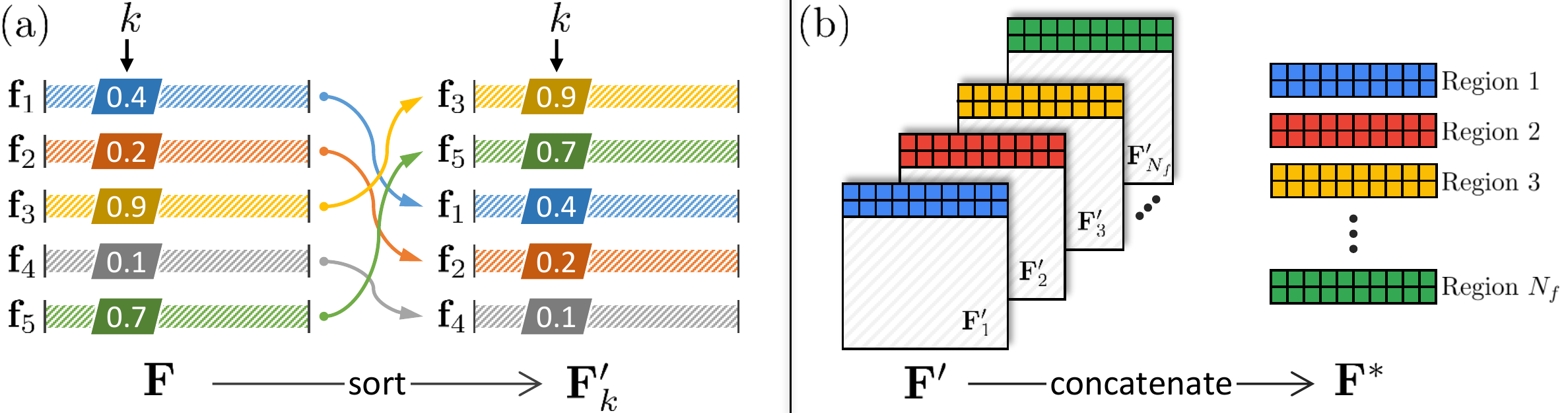}
\caption{Toy examples of (a) sorting the the $k$-th element of the vectors in the feature matrix $\mathbf{F}$ to build $\mathbf{F}'_k$ and consequently $\mathbf{F}'$ and (b) concatenation of the first $N_r$ rows of $\mathbf{F}'_k$ to construct the 2D matrix $\mathbf{F}^*$ which corresponds to the regions with high activations.}
\label{fig:softpool}
\end{figure}

The main challenge when processing a point cloud is its unstructured arrangement. This implies that changing the order of the points in $\mathbf{P}_\text{in}$ describes the same point cloud, but generates a different feature matrix that flows into our architecture with convolutional operators. 
To solve this problem, we propose to organize the feature vectors in $\mathbf{F}$ so that their $k$-th element are sorted in a descending order, which is denoted as $\mathbf{F}'_k$. Note that $k$ should not be larger than $N_f$.
A \emph{toy example} of this process is depicted in \figref{fig:softpool}(a) where we assume that there are only five points in the point cloud and arrange the five feature vectors from 
$\mathbf{F} = [\mathbf{f}_i]_{i = 1}^5$
to 
$\mathbf{F}'_k = [\mathbf{f}_i]_{i = \{3, 5, 1, 2, 4\}}$ 
by comparing the $k$-th element of each vector. 
Repeating this process for all the $N_f$ elements in $\mathbf{f}_i$, all $\mathbf{F}'_k$ together result to a 3D tensor $\mathbf{F}'=[\mathbf{F}'_1, \mathbf{F}'_2, \dots \mathbf{F}'_{N_f}]$ with the dimension of $N_\text{in} \times N_f\times N_f$. 
As a result, any permutation of the points in $\mathbf{P}_\text{in}$ generate the same $\mathbf{F}'$.

\begin{figure}[!b]
\centering
\includegraphics[width=1.0\linewidth]{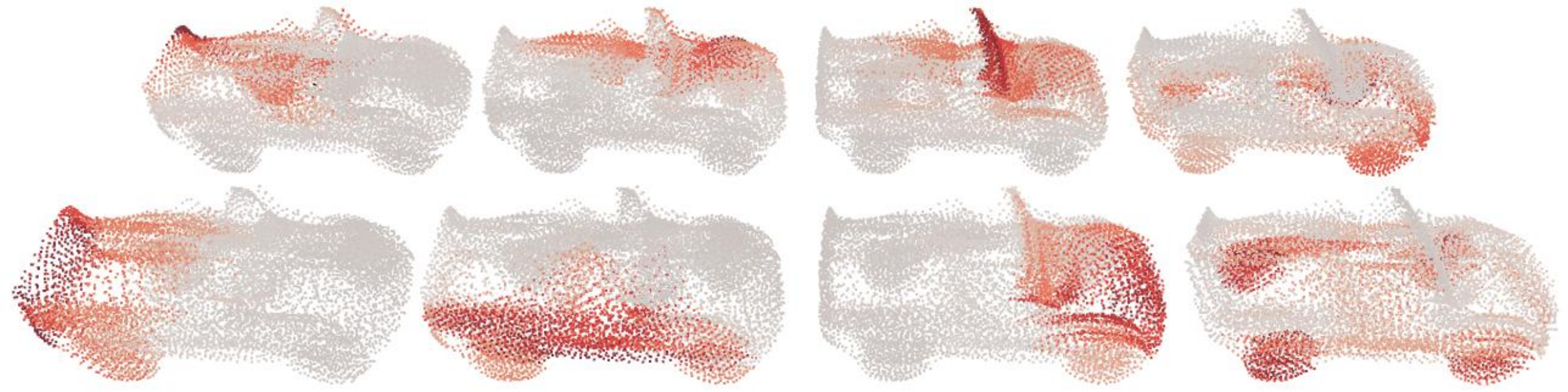}
\caption{Deconstructing the learned regions (unsupervised) that correspond to different parts of the car.} 
\label{fig:regions}
\end{figure}

Sorting the feature vectors in a descending order highlights the ones with the highest activation values. Thus, by selecting the first $N_r$ feature vectors from all the $\mathbf{F}'_k$ as shown in \figref{fig:softpool}(b), we assemble $\mathbf{F}^*$ that accumulates the features with the highest activations.
Altogether, the output of soft pooling is the $N_f \cdot N_r$ point features.
Since each feature vector corresponds to a point in $\mathbf{P}_\text{in}$, we can interpret the first $N_r$ feature vectors as a region in the point cloud.
The effects of the activations on the 3D reconstruction are illustrated in \figref{fig:regions}, where the point cloud is divided into $N_f$ regions.
Later in \secref{sec:regional_div}, we discuss on how to learn $\mathbf{W}_\text{point}$ by incorporating these regions. That section introduces several loss functions which optimize towards entropy, Chamfer distance and earth-moving distance such that each point is optimized to fall into only one region and to be selected for $\mathbf{F}^*$ by maximizing the $k$-th element of the feature vector associated to the same region.

\begin{figure}[!t]
\centering
\includegraphics[width=1.0\linewidth]{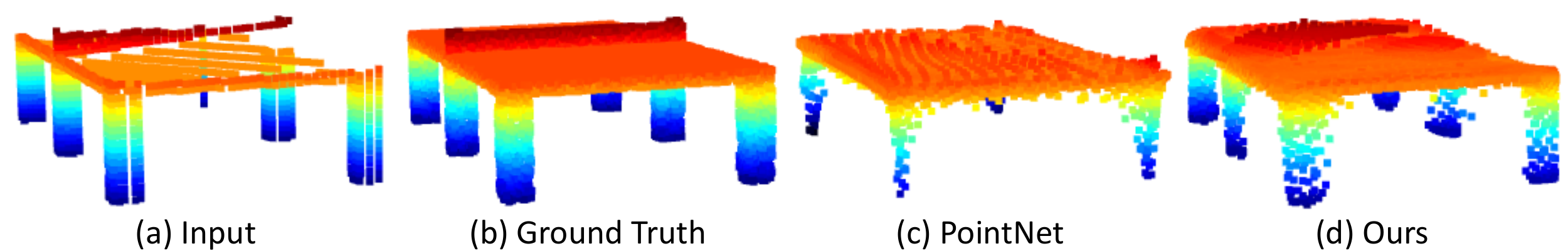}
\caption{Comparison of our method and PointNet~\cite{qi2017pointnet} where PointNet reconstructs the more typical four-leg table instead of six in (c).}
\label{fig:pointnet}
\end{figure}

Similar to PointNet~\cite{qi2017pointnet}, we also rely on MLP to build the feature matrix $\mathbf{F}$. 
However, PointNet directly applies max-pooling on $\mathbf{F}$ to produce a vector while we try to generalize this approach and sort the feature vectors in order to assemble a matrix $\mathbf{F}^*$ as illustrated in \figref{fig:softpool}.
Considering the distinction between the two approaches, we refer our approach as \emph{soft pooling}.
Fundamentally, in addition to the increased amount of information from our feature vectors, the advantage of our method is the ability to apply regional convolutional operations to $\mathbf{F}^*$, as discussed in \secref{sec:rconv}.
The differences are evident in \figref{fig:pointnet}, where the proposed method achieves detailed results on reconstructing all the six legs while PointNet follows the more generic structure of the table with four. This proves that \emph{soft pooling} makes our decoder able to take all observable geometries into account to complete the shape, while the max-pooled PointNet feature cannot reveal the rarely seen geometry.

\section{Regional convolution}
\label{sec:rconv}

Operating on $\mathbf{F}^*$, 
we introduce the convolutional kernel $\mathbf{W}_\text{conv}$ that transforms $\mathbf{F}^*$ to a new set of points $\mathbf{P}_\text{conv}$  by taking several point features into consideration. We structure $\mathbf{W}_\text{conv}$ with a dimension of $N_p \times N_f \times 3$ where $N_p$ represent the number of points which are taken into consideration such that 
\begin{align}
\mathbf{P}_\text{conv}(i,j) = 
\sum_{l=1}^{N_f} \sum_{k=1}^{N_p} 
\mathbf{F}^*(i+k,l) \mathbf{W}_\text{conv}(j,k,l)~.
\end{align}
Here, the kernel slides only inside each region of features without taking features from two different regions in one convolutional operation. As the kernel size allows it to cover $N_p$ features, we pad each region with $N_p - 1$ duplicated samples at the end of each region in order to keep the output resolution the same as $N_\text{in}$. Experimentally, we tried different numbers of $N_p$ ranging from 4 to 64 and evaluated that 32 generates the best results. 
Learning the values in $\mathbf{W}_\text{conv}$ is discussed in \secref{sec:regional_div}.

In addition, we use a convolution stride which is set as a value smaller than $N_p$ to change the output resolution in terms of the number of point features.
With a stride of $S$, we then take samples every $S$ point feature in $\mathbf{F}^*$.
Notice that, by using a stride which is smaller than 1, we can also upsample $\mathbf{F}^*$ by interpolating $\frac{1}{S}-1$ new points between two points then apply the convolution kernel again. This is an essential tool in reconstructing the object from a partial scan.

\section{Network architecture}

We build an encoder-decoder architecture which consists of MLP and our regional convolutions, respectively.  
Serving as the input to our network, we permutate the input scans and resample 1,024 points. If the partial scans have less than 1,024 points, we then duplicate the missing samples.

Our encoder is a point-wise MLP that generates the output neuron with a dimension of $[512, 512, 8]$. 
We then perform soft pooling as described in \secref{sec:soft_pooling} that produces $\mathbf{F}^*$ with the size of $[256, 8]$ by setting $N_r$ to 32 and $N_f$ to 8, resulting an output of $N_f \cdot N_r=256$ features.

\begin{figure}[!t]
\centering
\includegraphics[width=\linewidth]{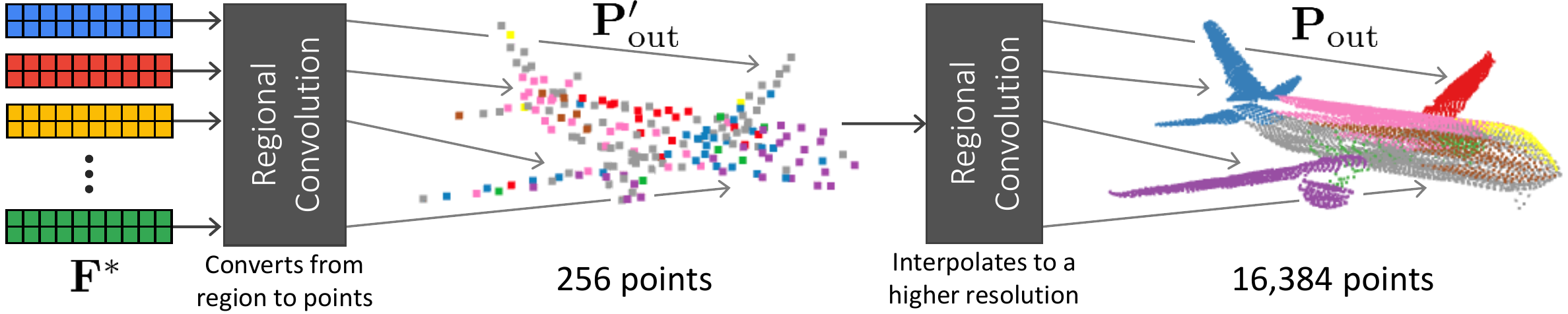}
\caption{Decoder architecture of SoftPoolNet with two regional convolution that converts the features from the regions to point clouds and interpolates from the coarse 256 points to a higher resolution with 16,384 points.}
\label{fig:regional_convolution}
\end{figure}

Finally, for the decoder, we propose a two-stage point cloud completion architecture which is trained end-to-end.
The output of the first is used as the input of the second point cloud completion network. 
Both of them produces the completed point cloud but with different resolutions. 
Illustrated in \figref{fig:regional_convolution},
we construct the decoder with two regional convolutions from \secref{sec:rconv}. 
The first output $\mathbf{P}_\text{out}'$ is fixed at 256 while the second $\mathbf{P}_\text{out}$ produces a maximum resolution of 16,384.

\section{Loss functions}
\label{sec:regional_div}

During learning, we evaluate whether the predicted point feature  $\mathbf{P}_\text{out}$ matches the given ground truth $\mathbf{P}_\text{gt}$ through the Chamfer distance. 
Similar to \cite{Groueix_2018_CVPR,yang2018foldingnet,yuan2018pcn}, 
we use the regression loss function for the shape completion from a point cloud
\begin{align}
\mathcal{L}_\text{complete}(\mathbf{W}_\text{point}, \mathbf{W}_\text{conv}) = \text{Chamfer}(\mathbf{P}_\text{out}, \mathbf{P}_\text{gt}) ~.
\end{align}
We observed that there are two major drawbacks in using this loss function alone -- the reconstructed surface tends to be either curved on the sharp edges such FoldingNet~\cite{yang2018foldingnet} or having noisy points appear on flat surfaces such as AtlasNet~\cite{Groueix_2018_CVPR} and PCN~\cite{yuan2018pcn}.
In this work, we tackle these problems by finding local regions first, then by optimizing the inter- and intra-regional relationships.

Moreover, while FoldingNet~\cite{yang2018foldingnet} sacrifices local details to present the entire model with a single mesh having smooth surface, AtlasNet~\cite{Groueix_2018_CVPR} and PCN~\cite{yuan2018pcn} use local regions (or patches) to increase the details in the 3D model. However, both of them~\cite{Groueix_2018_CVPR,yuan2018pcn} have severe overlapping effects between adjacent regions which makes the generated object noisy and the regions discontinuous.
To solve this problem, we aim at reducing the overlaps between two adjacent regions.



\subsection{Learning activations through regional entropy}
\label{sec:loss_activation}

Considering that the dimension of a single feature is $N_f$, we can directly define $N_f$ regions for all features.
Given the probabilities of regions to which the feature $\mathbf{f}_i$ belong, we want to optimize the inter- and intra-regional relationships among the features. We directly present the probability of the feature $\mathbf{f}_i$ belonging to all $N_f$ regions by applying the softmax function to $\mathbf{f}_i$ as
\begin{align}
	P(\mathbf{f}_k, i) = \frac{\mathbf{f}_k[i]}{\sum_{j=1}^{N_f}\mathbf{f}_k[j]} ~.
\end{align}
Since the information entropy evaluates both the distribution and the confidence of the probabilities of a set of data, we define the feature entropy  and the regional entropy  based on the regional probability of the feature.

The goal of the inter-regional loss function is to similarly distribute the number of points throughout the regions. We define the regional entropy as 
\begin{align}
	\mathcal{E}_r = 
	- \frac{1}{B} \sum_{j = 1}^{B} \sum_{i = 1}^{R} \left[
	\left(\frac{1}{N}\sum_{k=1}^{N}P(\mathbf{f}_k, i)\right) \cdot 
	\log{\left(\frac{1}{N}\sum_{k=1}^{N}P(\mathbf{f}_k, i)\right)} \right]
\end{align}
where $B$ is the batch-size. Here, we want to maximize $\mathcal{E}_r$. Considering that the upper-bound of $\mathcal{E}_r$ is $-\log\frac{1}{R} = \log(R)$, we can then define the inter-regional loss function as 
\begin{align}
\mathcal{L}_\text{inter}(\mathbf{W}_\text{point}) &=  \log (R) -\mathcal{E}_r
\end{align}
in order to acquire a positive loss function. Once $\mathcal{E}_r$ is close to $\log (R)$, each region would contain similar amount of point features.
Interestingly, we can select 
the number of regions by evaluating how much the regional entropy $\mathcal{E}_r$ differs from its upper-bound. 
The best number of regions should be the one with a small $\mathcal{L}_\text{inter}$. This is evaluated later in  \tabref{tab:shapenet_regions}.

On the other hand, the goal of the intra-regional loss function is to boost the confidence of each feature to be in a single region.
The intra-regional loss function then minimize the feature entropy 
\begin{align}
\mathcal{L}_\text{intra}(\mathbf{W}_\text{point}) =
	 - \frac{1}{N}\frac{1}{B} \sum_{k=1}^{N}\sum_{j = 1}^{B}\sum_{i = 1}^{N_f}P(\mathbf{f}_k, i) \log{P(\mathbf{f}_k, i)}~.
\end{align}
The optimum case of the feature entropy is for each feature to be a one-hot code, \ie when only one element is 1 while the others are zero.

\subsection{Reducing the overlapping regions}
\label{sec:loss_surface}

\begin{figure}[!t]
\centering
    \includegraphics[width=\linewidth]{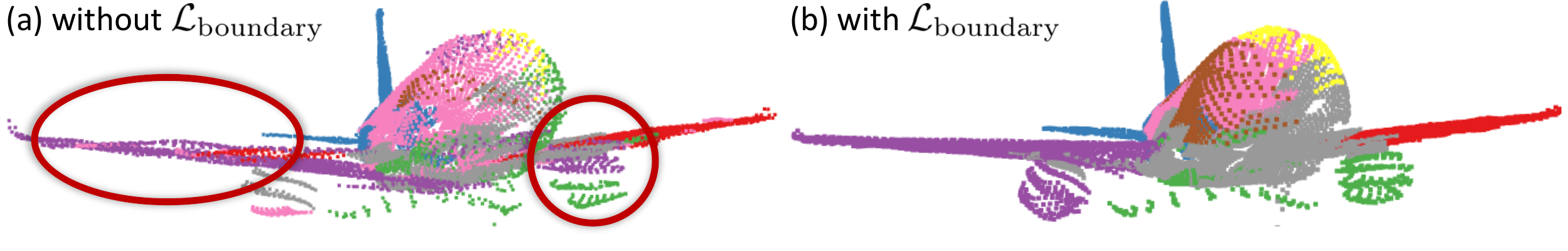}
  \caption{Effects of without and with $\mathcal{L}_\text{boundary}$  where the wings are not planar and the engines are less visible in (a). Note that the colors represent different regions. \label{fig:car}}
\end{figure}

Although $\mathcal{L}_\text{intra}$ tries to make each point feature  confident about the region to which it belongs, instances exist where many adjacent points would fall under different regions. For example, we observe in \figref{fig:car} that patches from different regions are stacked on top of each other, producing noisy reconstructions. 
Notably, this introduces unexpected results when fitting a mesh to the point cloud.
Thus, we want to minimize region overlap by optimizing the network to restrict the connection between adjacent regions to their boundaries.


First, each point is assigned to a region with the highest activation. All points that belong to region $i$ but has activation for region $j$ larger than a threshold $\tau$ are included in the set $\mathcal{B}_i^j$. Inversely, the points that belong to region $j$ but have activation for region $i$ larger than $\tau$ are added in the set $\mathcal{B}_j^i$. 
Note that, if both sets $\mathcal{B}_i^j$ and $\mathcal{B}_j^i$ are not empty, the regions $i$ and $j$ are then adjacent.
Thus, by minimizing the Chamfer distance between $\mathcal{B}_i^j$ and $\mathcal{B}_j^i$, we can make the overlapping sets of points smaller such that the optimal result is a line.
We then define the loss function for the boundary as
\begin{align}
\mathcal{L}_\text{boundary}(\mathbf{W}_\text{point}, \mathbf{W}_\text{conv}) = \sum_{i = 1}^{N_f} \sum_{j = i}^{N_f} \text{Chamfer}(\mathcal{B}_i^j, \mathcal{B}_j^i)
\end{align}
where both $\mathbf{W}_\text{point}$ and $\mathbf{W}_\text{conv}$ are optimized.
After experimenting on different values of $\tau$ from 0.1 to 0.9, we set $\tau$ to be 0.3.

\begin{figure}[!t]
  \centering
    \includegraphics[width=\linewidth]{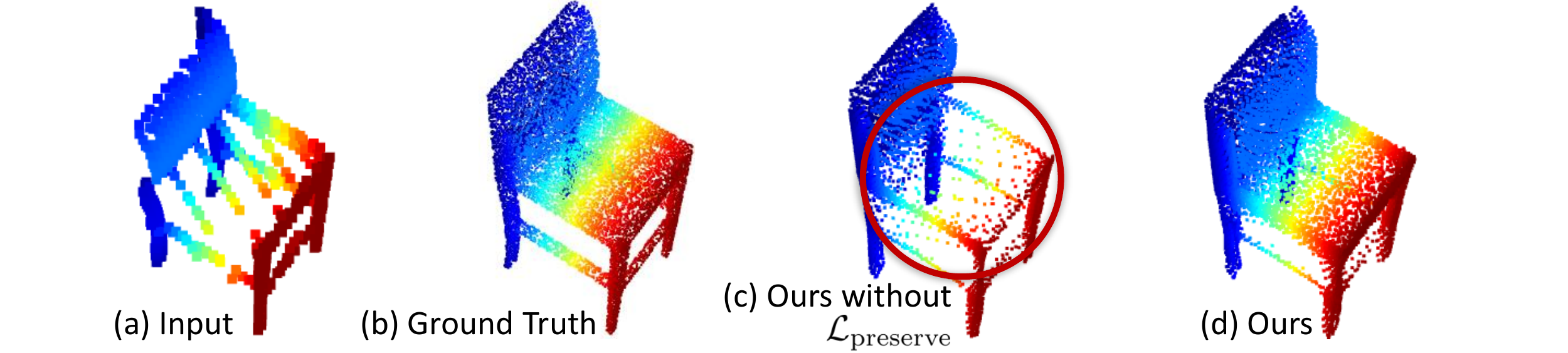}
  \caption{Effects of without and with $\mathcal{L}_\text{preserve}$ where the seat is missing in (c).}
  \label{fig:preserve}
\end{figure}

\subsection{Preserving the features from MLP}

After sorting and filtering the features to produce $\mathbf{F}^*$, some feature vectors in $\mathbf{F}^*$ are duplicated while some vectors from $\mathbf{F}$ are missing in $\mathbf{F}^*$.
To avoid these, we introduce the loss function
\begin{align}
\mathcal{L}_\text{preserve}(\mathbf{W}_\text{point}) = \text{Earth-moving}(\mathbf{F}^*, \mathbf{F}) ~.
\end{align}  
Since the earth moving distance~\cite{Li_2013_ICCV} is not efficient when the size of the samples is large, we then randomly select 256 vectors from $\mathbf{F}$ and $\mathbf{F}^*$. Considering that the feature dimensions in $\mathbf{F}$ and $\mathbf{F}^*$ are both $N_f$, the earth moving distance then takes features with $N_f$ dimension as input.
%
In practice, \figref{fig:preserve} visualizes the effects of $\mathcal{L}_\text{preserve}$ in the reconstruction, where removing this loss produce a large hole on the seat while incorporating this loss builds a well-distributed point cloud.

\section{Experiments}
\label{sec:experiments}

For all evaluations, we train our model with an NVIDIA Titan V and parameterize it with a batch size of 8. 
Moreover, we apply the Leaky ReLU with a negative slope of 0.2 on the output of each regional convolution output.

\subsection{Object completion on ShapeNet}
\label{sec:shapenet}

We evaluate the performance of the geometric completion of a single object on the ShapeNet~\cite{chang2015shapenet} database where training data are paired point clouds of the partial scanning and the completed shape. To make it comparable to other approaches, we adopt the standard 8 category evaluation~\cite{yuan2018pcn} for a single object completion. 
As rotation errors are common in the partial scans, we further evaluate our approach against other works on the ShapeNet database with rotations. We also evaluate the performance on both high and low resolutions which contain 16,384 and 2,048 points, respectively.

We compare against other point cloud completion approaches such as PCN~\cite{yuan2018pcn}, FoldingNet~\cite{yang2018foldingnet}, AtlasNet~\cite{Groueix_2018_CVPR} and PointNet++~\cite{qi2017pointnet++}. To show the advantages over volumetric completion, we also compare against 3D-EPN~\cite{yang2018dense} and ForkNet~\cite{wang2019forknet} with an output resolution of $64 \times 64 \times 64$.
Notably, we achieve the best results on most objects and in all types of evaluations as presented in \tabref{tab:shapenet_cd_16384}, \tabref{tab:shapenet_cd_2048} and \tabref{tab:shapenet_emd}.

\begin{table*}[!b]
\centering
\begin{tabular}{l|cccccccc|c}
\multicolumn{3}{l}{Output Resolution = 16,384} \\
\toprule	
 \multicolumn{1}{c}{Method} 
 & plane & cabinet & car & chair & lamp & sofa & table & vessel & \emph{Avg.} \\
\midrule 
       3D-EPN~\cite{dai2017shape} & 13.16 & 21.80 & 20.31 & 18.81 & 25.75 & 21.09 & 21.72 & 18.54 & 20.15 \\
       ForkNet~\cite{wang2019forknet} & 9.08 & 14.22 & 11.65 & 12.18 & 17.24 & 14.22 & 11.51 & 12.66 & 12.85 \\
       PointNet++~\cite{qi2017pointnet++} & 10.30 & 14.74 & 12.19 & 15.78 & 17.62 & 16.18 & 11.68 & 13.52 & 14.00 \\
       FoldingNet~\cite{yang2018foldingnet} & 5.97 & 10.80 & 9.27 & 11.25 & 12.17 & 11.63 & 9.45 & 10.03 & 10.07 \\
       \with{FoldingNet}{$\mathcal{L}_\text{boundary}$} & 5.79 & 10.61 & 8.62 & 10.33 & 11.56 & 11.05 & 9.41 & 9.79 & 9.65 \\
       PCN~\cite{yuan2018pcn} & 5.50 & 10.63 & 8.70 & 11.00 & 11.34 & 11.68 & 8.59 & 9.67 & 9.64 \\
       \with{PCN}{$\mathcal{L}_\text{boundary}$} & 5.13 & 9.12 & 7.58 & 9.35 & 9.40 & 9.31 & 7.30 & 8.91 & 8.26 \\
       \textbf{Our Method} & \textbf{4.01} & \textbf{6.23} & \textbf{5.94} & \textbf{6.81} & \textbf{7.03} & \textbf{6.99} & \textbf{4.84} & \textbf{5.70} & \textbf{5.94} \\ 
\bottomrule
\end{tabular}
\caption{Completion evaluated by means of the Chamfer distance  (multiplied by $10^3$) with the output resolution of 16,384. 
\label{tab:shapenet_cd_16384}
}
\end{table*}

\begin{table*}[!t]
\centering
\begin{tabular}{l|cccccccc|c}
\multicolumn{3}{l}{Output Resolution = 2,048} \\
\toprule	
 \multicolumn{1}{c}{Method} 
  & plane & cabinet & car & chair & lamp & sofa & table & vessel & \emph{Avg.} \\
\midrule 
       FoldingNet~\cite{yang2018foldingnet} & 11.18 & 20.15 & 13.25 & 21.48 & 18.19 & 19.09 & 17.80 & 10.69 & 16.48 \\
       \with{FoldingNet}{$\mathcal{L}_\text{boundary}$} & 11.09 & 19.95 & 13.11 & 21.27 & 18.22 & 19.06 & 17.62 & 10.10 & 16.30 \\
       AtlasNet~\cite{Groueix_2018_CVPR} & 10.37 & 23.40 & 13.41 & 24.16 & 20.24 & 20.82 & 17.52 & 11.62 & 17.69 \\
       \with{AtlasNet}{$\mathcal{L}_\text{boundary}$} & 9.25 & 22.51 & 12.12 & 22.64 & 18.82 & 19.11 & 16.50 & 11.53 & 16.56 \\
       PCN~\cite{yuan2018pcn} & 8.09 & 18.32 & 10.53 & 19.33 & 18.52 & 16.44 & 16.34 & 10.21 & 14.72 \\
       \with{PCN}{$\mathcal{L}_\text{boundary}$} & 6.39 & 16.32 & 9.30 & 18.61 & 16.72 & 16.28 & 15.29 & 9.00 & 13.49 \\
       TopNet~\cite{tchapmi2019topnet} & 5.50 &12.02 & 8.90 & 12.56 & 9.54 & 12.20 & 9.57 & 7.51 & 9.72 \\
       \textbf{Our Method} & \textbf{4.76} & \textbf{10.29} & \textbf{7.63} & \textbf{11.23} & \textbf{8.97} & \textbf{10.08} & \textbf{7.13} & \textbf{6.38} & \textbf{8.31} \\
       \without{$\mathcal{L}_\text{inter}$} & 10.82 & 20.45 & 15.21 & 20.19 & 18.05 & 18.58 & 15.65 & 8.81 & 15.97 \\
       \without{$\mathcal{L}_\text{intra}$} & 5.23 & 16.10 & 12.49 & 14.62 & 13.90 & 12.37 & 12.96 & 5.72 & 11.67 \\
       \without{$\mathcal{L}_\text{inter}$, $\mathcal{L}_\text{intra}$} & 10.91 & 20.54 & 15.27 & 20.28 & 18.16 & 18.66 & 15.75 &  8.91 & 16.06 \\
       \without{$\mathcal{L}_\text{boundary}$} & 5.46 & 10.98 &  8.27 & 11.95 &  9.51 & 10.92 &  7.78 &  7.40 & 9.03 \\
       \without{$\mathcal{L}_\text{preserve}$} & 10.29 & 19.75 & 14.13 & 19.35 & 17.88 & 18.21 & 15.23 & 8.11 & 15.37 \\
\bottomrule
\end{tabular}
\caption{Completion evaluated using the Chamfer distance (multiplied by $10^3$) with the output resolution of 2,048. 
}
\label{tab:shapenet_cd_2048}
\end{table*}


An interesting hypothesis is the capacity of $\mathcal{L}_\text{boundary}$ to be integrated in other existing approaches. Thus, \tabref{tab:shapenet_cd_16384} and \tabref{tab:shapenet_cd_2048} also evaluate this hypothesis and prove that this activation helps FoldingNet~\cite{yang2018foldingnet}, PCN~\cite{yuan2018pcn} and AtlasNet~\cite{Groueix_2018_CVPR} perform better.
Nevertheless, even with such improvements, the complete version of the proposed method still outperforms them.

\begin{table*}[!t]
\centering
\begin{tabular}{l|cccccccc|c}
	\multicolumn{3}{l}{Output Resolution = 1,024} \\
	\toprule	
	\multicolumn{1}{c}{Method} & ~plane~ & ~cabinet~ & ~car~ & ~chair~ & ~lamp~ & ~sofa~ & ~table~ & ~vessel~ & ~\emph{Avg.}~ \\
	\midrule 
	3D-EPN~\cite{dai2017shape} & 6.20 & 7.76 & 8.70 & 7.68 & 10.73 & 8.08 & 8.10 & 8.17 & 8.18 \\
	PointNet++~\cite{qi2017pointnet++} & 5.96 & 11.62 & 6.69 & 11.06 & 18.58 & 10.26 & 8.61 & 8.38 & 10.14 \\
	FoldingNet~\cite{yang2018foldingnet} & 15.64 & 22.13 & 17.46 & 29.74 & 32.00 & 24.57 & 18.99 & 21.88 & 22.80 \\
	PCN~\cite{yuan2018pcn} & 3.88 & 7.07 & 5.50 & 6.81 & 8.46 & 7.24 & 6.01 & 6.27 & 6.40 \\
	DeepSDF~\cite{park2019deepsdf} & 3.88 & - & - & 5.63 & - & \textbf{4.68} & - & - & -\\
	LGAN~\cite{pmlr-v80-achlioptas18a} & 3.32 & - & - & 5.59 & - & - & - & - & - \\
	MAP-VAE~\cite{Han_2019_ICCV} & 3.23 & - & - & 5.57 & - & - & - & - & - \\
	\textbf{Our Method} & \textbf{2.52} & \textbf{5.49} & \textbf{4.08} & \textbf{5.20} & \textbf{6.17} & 5.25 & \textbf{4.61} & \textbf{5.80} & \textbf{4.89} \\
	\bottomrule
\end{tabular}
\caption{Completion results using the Earth-Moving distance (multiplied by $10^2$) with the output resolution of 1,024. We report the values of DeepSDF~\cite{park2019deepsdf} from their original paper by rescaling according to the difference of point density. 
\label{tab:shapenet_emd}
}
\end{table*}

\begin{table}[!t]
\centering
\begin{tabular}{l|ccc}
	\toprule	
		\multicolumn{1}{c}{Method} 
	 & ~~~Fidelity~~~ & ~~~MMD~~~ & ~Consistency~ \\
	\midrule
		FoldingNet~\cite{yang2018foldingnet} & 0.03155 & 0.02080 & 0.01326 \\
		AtlasNet~\cite{Groueix_2018_CVPR} & 0.03461 & 0.02205 & 0.01646 \\
		PCN~\cite{yuan2018pcn} & 0.02800 & 0.01850 & 0.01163 \\
		\textbf{Our Method} & \textbf{0.02171} & \textbf{0.01465} & \textbf{0.00922} \\ 
	\midrule
		PCN~\cite{yuan2018pcn} (rotate) & 0.03352 & 0.02370 & 0.01639  \\
		\textbf{Our Method} (rotate)~ & \textbf{0.02392} & \textbf{0.01732} & \textbf{0.01175} \\ 
	\bottomrule
	\end{tabular}
\caption{Car completion on LiDAR scans from KITTI. }
\label{tab:kitti_completion}
\end{table}

\subsection{Car completion on KITTI}

The KITTI~\cite{geiger2013vision} dataset present partial scans of real-world cars using Velodyne 3D laser scanner.
We adopt the same training and validating procedure for car completion as proposed by PCN~\cite{yuan2018pcn}. 
We train a car completion model based on the training data generated from ShapeNet~\cite{chang2015shapenet} and test our completion method on sparse point clouds generated from the real-world LiDAR scans. 
For each sample, the points within the bounding boxes are extracted with 2,483 partial point clouds. Each point cloud is then transformed to the box's coordinates to be completed by our model then transformed back to the world frame.
PCN~\cite{yuan2018pcn} proposed three metrics to evaluate the performance of our model:
(1)~\emph{Fidelity}, \ie the average distance from each point in the input to its nearest neighbour in the output (\ie measures how well the input is preserved);
(2)~\emph{Minimal Matching Distance} (MMD), \ie the Chamfer distance between the output and the car's point cloud nearest neighbor from ShapeNet 
(\ie measures how much the output resembles a typical car); and, 
(3)~\emph{Consistency}, the average Chamfer distance between the completion outputs of the same instance in consecutive frames (\ie measures how consistent the network’s outputs are against variations in the inputs).

\tabref{tab:kitti_completion} shows that we achieve state of the art on the metrics compared to FoldingNet~\cite{yang2018foldingnet}, 
AtlasNet~\cite{Groueix_2018_CVPR} and 
PCN~\cite{yuan2018pcn}.
When we introduce random rotations on the bounding box in order to simulate errors in the initial stages, we still acquire the lowest errors.

\subsection{Classification on ModelNet and PartNet}

We evaluate the performance of the features in term of classification on ModelNet10~\cite{7298801}, ModelNet40~\cite{7298801} and PartNet~\cite{Mo_2019_CVPR} datasets.
ModelNet40 contains 12,311 CAD models in 40 categories. 
Here, the training data contains 9,843 samples and the testing data contains 2,468 samples.
Following RS-DGCNN~\cite{sauder2019self}, a linear Support Vector Machine~\cite{cortes1995support} (SVM) is trained on the representations learned in an unsupervised manner on the ShapeNet dataset. 
RS-DGCNN~\cite{sauder2019self} divides the point cloud of the objects into several regions by positioning the object in a pre-defined voxel grid, then use the regional information to help train latent feature. 
In \tabref{tab:modelnet}, the proposed method outperforms RS-DGCNN~\cite{sauder2019self} by 1.64\% accurracy on ModelNet40 dataset, which shows that our feature contains better categorical information. Notably, similar results are also acquired from ModelNet10~\cite{7298801} and PartNet~\cite{Mo_2019_CVPR} with their respective evaluation strategy.

\begin{table}[!t]
 \centering
 \begin{tabular}{l|ccc}
 	\toprule	
 		\multicolumn{1}{c}{Method} 
 		  & ~ModelNet40~\cite{7298801}~ & ~ModelNet10~\cite{7298801}~ & ~PartNet~\cite{Mo_2019_CVPR}~\\
 	 \midrule
 	     VConv-DAE & 75.50\% & 80.50\% & - \\
         3D-GAN & 83.30\% & 91.00\% & 74.23\% \\
         Latent-GAN & 85.70\% & 95.30\% & - \\
         FoldingNet & 88.40\% & 94.40\% & - \\
         VIP-GAN & 90.19\% & 92.18\% & - \\
         RS-PointNet~\cite{sauder2019self} & 87.31\% & 91.61\% & 76.95\% \\
         RS-DGCNN~\cite{sauder2019self} & 90.64\% & 94.52\% & - \\
         KCNet~\cite{shen2018mining} & 91.0\% & 94.4\% & - \\
 	\midrule
 		\textbf{Our Method} & \textbf{92.28\%} & \textbf{96.14\%} & \textbf{84.32\%}\\
 		\without{$\mathcal{L}_\text{inter}$} & 89.40\% & 95.75\%  & 81.13\% \\
 		\without{$\mathcal{L}_\text{intra}$} & 83.70\% & 90.21\%  & 79.28\%\\
        \without{$\mathcal{L}_\text{inter}$, $\mathcal{L}_\text{intra}$}~ & 82.97\% & 90.02\% & 78.41\% \\
 		\without{$\mathcal{L}_\text{boundary}$} & 88.26\% & 95.01\% & 80.86\% \\
 		\without{$\mathcal{L}_\text{preserve}$} & 86.09\% & 92.27\% & 79.05\% \\
 	\bottomrule
 	\end{tabular}
 \caption{Object classification on ModelNet40~\cite{7298801}, ModelNet40~\cite{7298801} and PartNet~\cite{Mo_2019_CVPR} datasets in terms of accuracy.
 \label{tab:modelnet}
 }
 \end{table}

\begin{table*}[!b]
\centering
\begin{tabular}{l|ccccc}
    \toprule
    $(N_f, N_r)$ & ~~$(2, 128)$~~ & ~~$(4, 64)$~~ & ~~$(8, 32)$~~ & ~~$(16, 16)$~~ & ~~$(32, 8)$~~ \\
    \midrule
    Chamfer Distance~ & 7.80 & 6.31 & \textbf{5.94} & 6.27 & 6.75 \\
    $\mathcal{L}_\text{inter}$ & 0.41 & 0.67 & \textbf{0.20} & 0.49 & 1.33 \\
    \bottomrule
\end{tabular}
\caption{Influence of  $N_f$ and  $N_r$ on the Chamfer distance (multiplied by $10^3$) and $\mathcal{L}_\text{inter}$.
}
\label{tab:shapenet_regions}
\end{table*}

\subsection{Ablation study}
 
\paragraph{Loss functions.} In the reconstruction and classification experiments, \tabref{tab:shapenet_cd_2048} and \tabref{tab:modelnet} also include the ablation study that investigates the effects of the loss functions from \secref{sec:regional_div}. For both experiments, we notice all loss functions are critical to achieve good results since each of them focuses on different aspects.

\paragraph{Activations.}
Since the number of regions is one of the hyper-parameters in our approach, we evaluate on the performance with different number of regions quantitatively in \tabref{tab:shapenet_regions}.
These results demonstrate that the accuracy for the shape completion is increasing as the number of regions increases from 2 to 8, then the performance gradually drops as the number of regions continues to increase from 8 to 32. By observing $\mathcal{L}_\text{inter}$ at the same time, we find that it achieves the minimum value of 0.20 when there are 8 regions as well. This proves that $\mathcal{L}_\text{inter}$ can be used as an indicator for whether the expected number of regions could be used or not.

\begin{figure}[!t]
\centering
\includegraphics[width=1.0\linewidth]{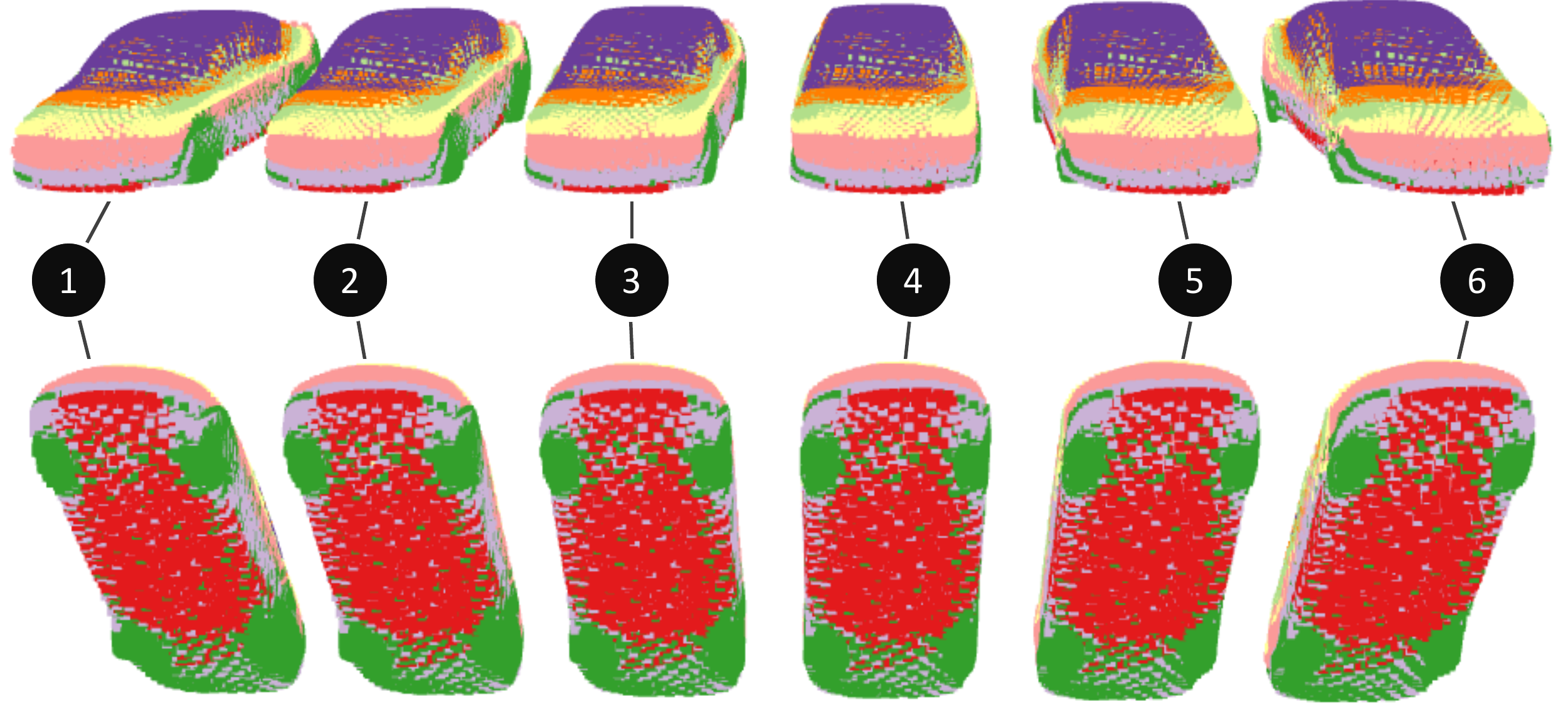}
\caption{With identical results, this evaluation shows the robustness of the reconstruction when we randomly shuffle the input point cloud.}
\label{fig:permute}
\end{figure}

\begin{figure*}[!t]
\centering
\includegraphics[width=\linewidth]{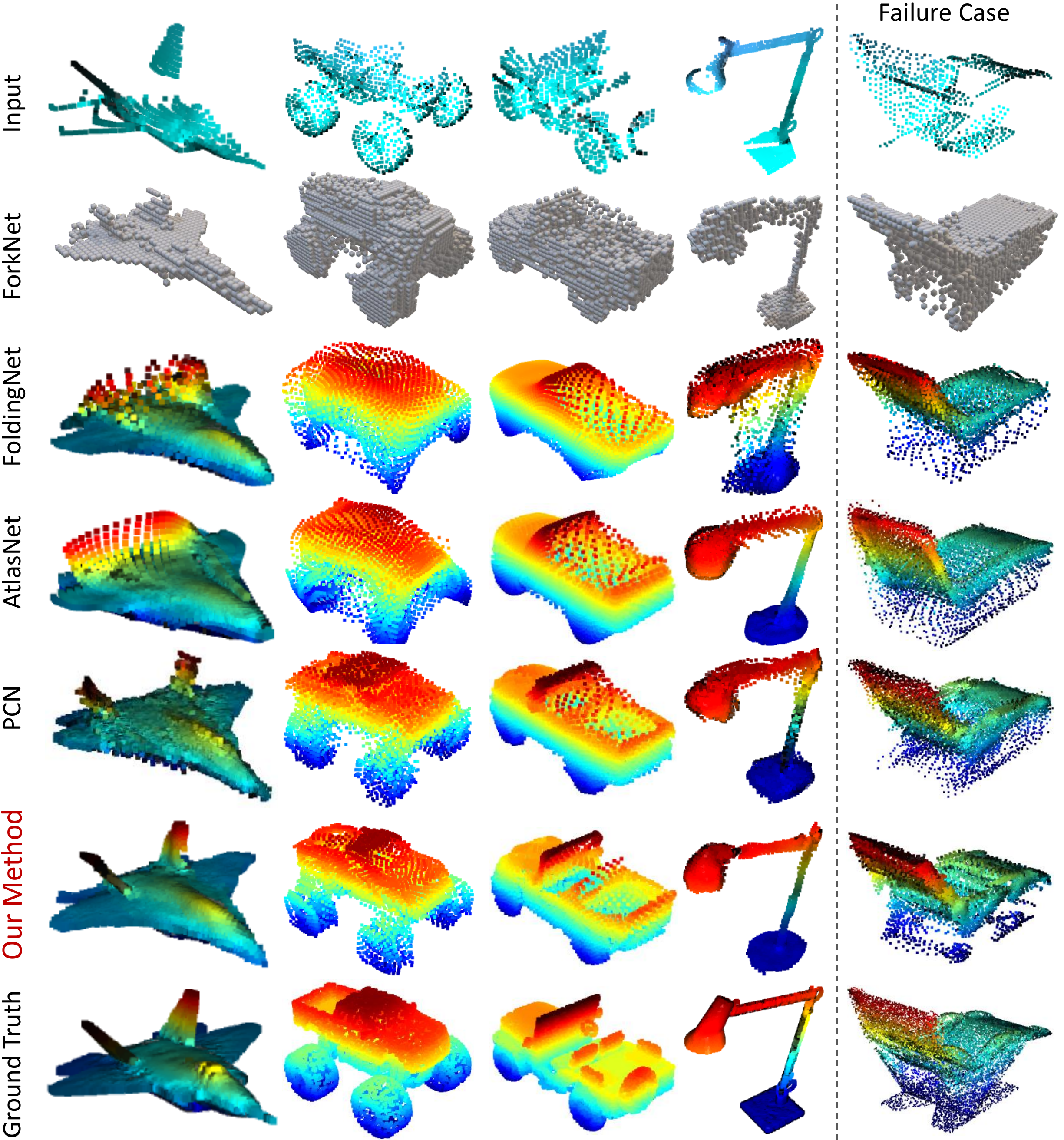}
\caption{Evaluated on ShapeNet~\cite{chang2015shapenet}, comparison of shape completion based on ForkNet~\cite{wang2019forknet}, FoldingNet~\cite{yang2018foldingnet}, AtlasNet~\cite{Groueix_2018_CVPR} and PCN~\cite{yuan2018pcn} against our method.}
\label{fig:qualitative}
\end{figure*}

\begin{table}[!b]
\centering
\begin{tabular}{l|c|c|c|c}
	\toprule	
		\multicolumn{1}{c}{Method} 
		 & \multicolumn{1}{c}{~~Size (MB)~~} & \multicolumn{1}{c}{~~Inference Time (s)~~} & \multicolumn{1}{c}{~~Closed Surface~~} & \multicolumn{1}{c}{~~Type of Data~~}\\
	\midrule
		3D-EPN~\cite{dai2017shape} & 420 & - & Yes & Volumetric \\
		ForkNet~\cite{wang2019forknet} & 362 & - & Yes & Volumetric \\
		FoldingNet~\cite{yang2018foldingnet} & 19.2 & 0.05 & Yes & Points \\
		AtlasNet~\cite{Groueix_2018_CVPR} & 2 & 0.32 & No & Points \\
		PCN~\cite{yuan2018pcn} & 54.8 & 0.11 & No & Points \\
		DeepSDF~\cite{park2019deepsdf} & 7.4 & 9.72 & Yes & SDF \\
		\textbf{Our Method} & 37.2 & 0.04 & Yes & Points \\
	\bottomrule
	\end{tabular}

\caption{Overview of the object completion methods. The inference time is the amount of time to conduct inference on a single sample.
\label{tab:param}
}
\end{table}

Moreover, \figref{fig:permute} shows the regional activations when we shuffle the sequence of points in the partial scan. We can see that both the reconstructed geometry relative sub-regions are identical. So, it illustrates that, by using the proposed regional activations, our model is permutation invariant, which indicates that the reordered point cloud is suitable to perform convolutions.

\paragraph{Point cloud versus volumetric data.}
In addition to achieving worse numerical results in \secref{sec:shapenet}, volumetric approaches have smaller resolutions than the point cloud approaches due to the memory constraints.
The difference becomes more evident in \figref{fig:qualitative}, where ForkNet~\cite{wang2019forknet} is limited by a $64 \times 64 \times 64$ grid.
Nevertheless, both the volumetric and point cloud approaches have difficulty in reconstructing thin structures. For instance, the volumetric approach tends to ignore the joints between the wheels and car chassis in \figref{fig:qualitative} while FoldingNet~\cite{yang2018foldingnet} and AtlasNet~\cite{Groueix_2018_CVPR} only use large surface to cover the area of wheels. In contrast, our approach is capable of reconstructing the thin structures quite well.
Moreover, in \tabref{tab:param}, we also achieve the lowest inference time compared to all point cloud and volumetric approaches.

\section{Conclusion}

This paper introduced the SoftPool idea as a novel and general way to extract rich deep features from unordered point sets such as 3D point clouds. 
Also, it proposed a state-of-the-art point cloud completion approach by designing a regional convolution network for the decoding stage. 
Our numerical evaluation reflects that our approach achieves the best results on different 3D tasks, while our quantitative results illustrate the reconstruction and completion ability of our method with respect to ground truth.

%
%
\bibliographystyle{splncs04}
\bibliography{egbib}
\end{document}


\pagestyle{headings}
\mainmatter
\def\ECCVSubNumber{5457}  

\title{SoftPoolNet: Shape Descriptor for Point Cloud Completion and Classification} 

\titlerunning{SoftPoolNet}
%
\author{Yida Wang\inst{1} \and
David Joseph Tan\inst{2} \and
Nassir Navab\inst{1} \and
Federico Tombari\inst{1,2}}
\authorrunning{Y. Wang et al.}
%
\institute{Technische Universit\"at M\"unchen\and
Google Inc.}

\maketitle


\section{Comparison of SoftPoolNet to PointNet and PCN}

Our architecture is composed by two parts: \emph{encoder} and \emph{decoder}.
The encoder takes the partial scan as input. We process the partial scans with our novel soft pooling to produce the ordered feature $F^*$.
Then, the decoder takes the feature $F^*$ as input and apply our regional convolution twice to produce the point clouds with resolutions of 256 and 16,384 successively.


Notably, there are some similar components between our encoder and PointNet~\cite{qi2017pointnet}, as well as our decoder and PCN~\cite{yuan2018pcn}. The following sections discuss the distinction in more detail.

\subsection{Distinction of our encoder from PointNet}

\begin{figure}[!h]
  \begin{minipage}[c]{0.99\linewidth}
    \includegraphics[width=\linewidth]{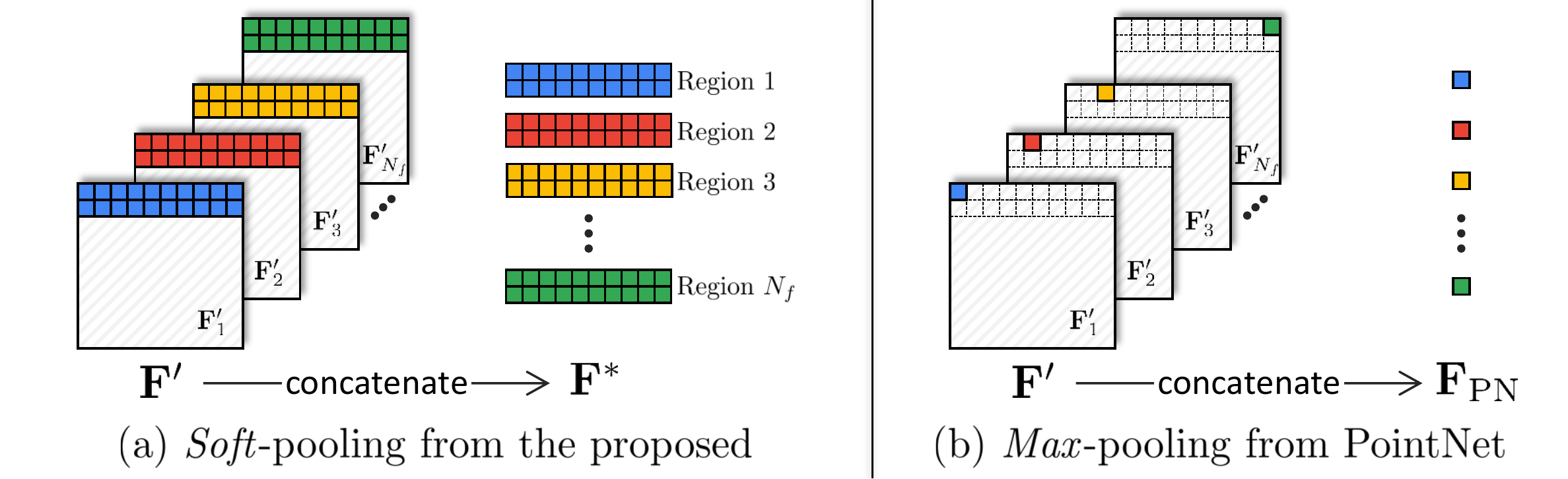}
  \end{minipage}
    \caption{Comparison between (a)~our soft-pool operation and (b)~max-pooling from PointNet, where the feature from PointNet is only a subset of our feature.}
    \label{fig:pointnet}
\end{figure}

Each point on the cloud goes through the multi-layer perceptron (MLP) to accumulate the feature vectors into the matrix $\mathbf{F}$.
Then, we sort the feature vectors in a descending order based on the $k$-th element of each vector. The sorted matrix is denoted as $\mathbf{F}'_{i}$. After independently sorting all $N_f$ elements, we collect the matrices to form the tensor $\mathbf{F}'$ as shown in \figref{fig:pointnet}(a).
We then build our softpool feature by taking the first $N_r$ elements of each matrix and concatenate them to $\mathbf{F}^*$.

When comparing our softpool feature $\mathbf{F}^*$ with the feature from PointNet~\cite{qi2017pointnet} denoted as $\mathbf{F}_\text{PN}$, PointNet executes a max-pooling operation on $\mathbf{F}'$ as illustrated in \figref{fig:pointnet}(b).
Assuming that both features are derived from the same $\mathbf{F}'$ produced by an MLP, we can conclude that $\mathbf{F}_\text{PN}$ is a subset of our feature where 
\begin{align}
\mathbf{F}_\text{PN} = 
\left[
\mathbf{F}'_{1}[1],~ \mathbf{F}'_{2}[2],~ \mathbf{F}'_{3}[3],~ \dots~ \mathbf{F}'_{N_f}[N_f]
\right]
\end{align}  
only takes the one value of each matrix while our method takes the first $N_r$ rows.
Due to this, the dimensionality of the feature are then distinct. PointNet takes a vector with 1,024 values while we take $N_r \times N_f \times N_f$.

Notably, both our softpool feature and the PointNet feature are permutation invariant, which means that $\mathbf{F}^*$ and $\mathbf{F}_\text{PN}$ are the same irrelevant of the order of the input points.
This is one of the most important aspect when handling point clouds since this data is unordered.




\subsection{Distinction of our decoder from PCN}

Based on our decoder architecture in the paper, the resulting feature from the encoder undergoes two successive regional convolution operations. The first converts the features to a course point cloud $\mathbf{P}'_\text{out}$ with 256 points. From there, the second regional convolution interpolates from the coarse to a fully-packed point cloud with 16,384 points which is denoted as $\mathbf{P}_\text{out}$.


Compared to PCN~\cite{yuan2018pcn}, both approaches execute a coarse-to-fine approach which is performed by our second regional convolution. 
However, the architecture and the method are different. 

Given $\mathbf{P}'_\text{out}$, PCN~\cite{yuan2018pcn} duplicates $\mathbf{P}'_\text{out}$ 64 times and appends a 2D coordinates of an $8 \times 8$ grid. 
Then, they use MLP to produce $\mathbf{P}_\text{out}$ that locally deforms the 2D grids around each point similar FoldingNet~\cite{yang2018foldingnet}.
In contrast, we interpolate 63 samples between every 2 points of $\mathbf{P}'_{\text{out}}$ and use the proposed regional convolution to produce $\mathbf{P}_{\text{out}}$. Compared to MLP in PCN, our regional convolution takes more local samples into account to produce a point in the higher resolution.




\section{Ablation study on the softpool feature $\mathbf{F}^*$}

\begin{figure}[!t]
  \centering
    \includegraphics[width=\linewidth]{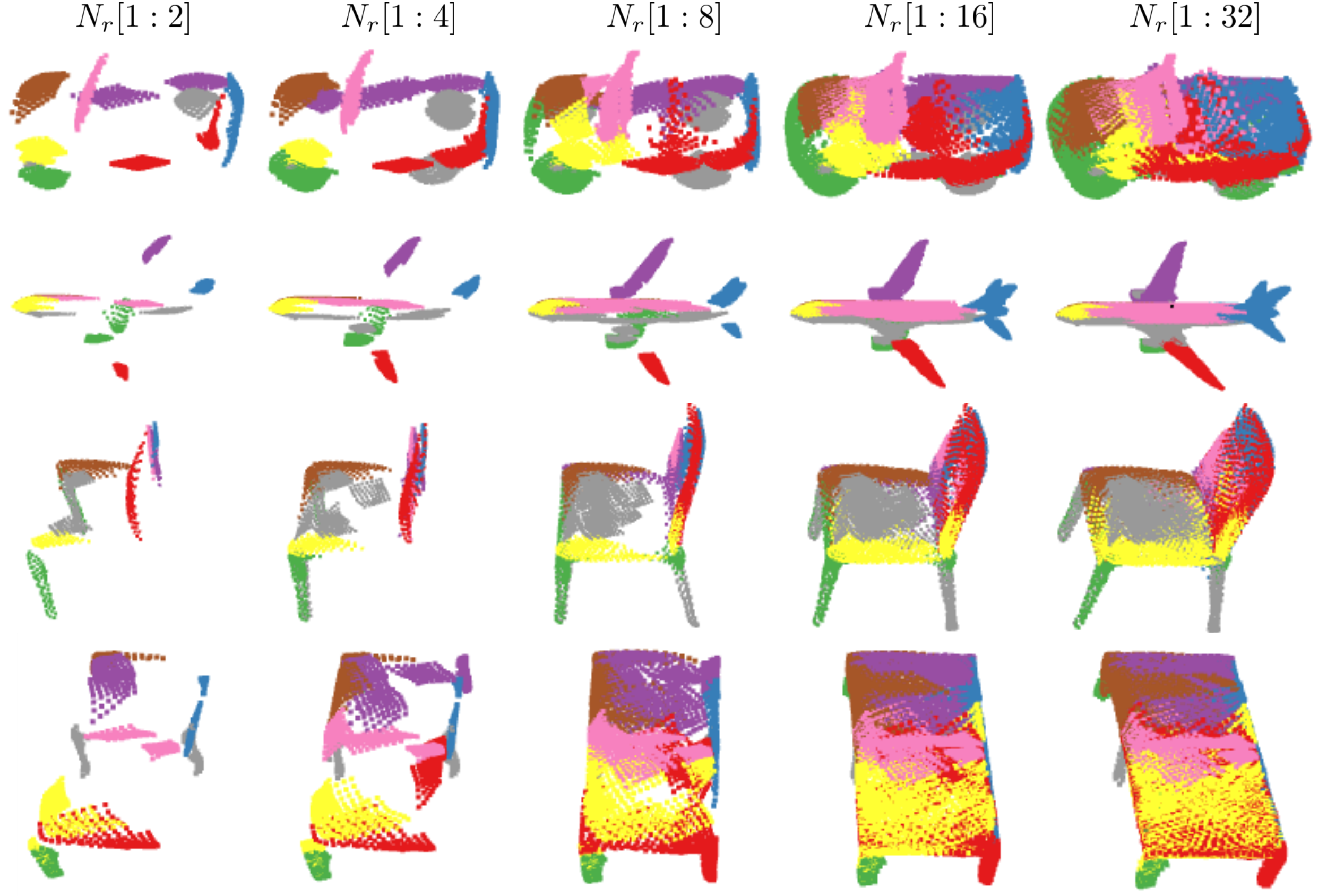}
    \caption{Results when choosing the first subsets of $N_r$ with the following ranges: $[1:2]$, $[1:4]$, $[1:8]$, $[1:16]$ and $[1:32]$ when the architecture is trained with $N_r=32$.}
    \label{fig:softpool_subsets}
\end{figure}

Using our architecture trained with $N_r =$ 32, we present the qualitative results when only a subset of the rows is selected.
The objective is to investigate which parts of the object each region reconstructs first.
In \figref{fig:softpool_subsets}, we start by limiting with the first two rows of the feature matrix then increasing $N_r$ to reach 32.
By selecting the first 2 features, we observe that the softpool feature focuses on a skeleton of the object without large surfaces. 
Although the regions reconstruct different parts of the object, they tend to cover the important components like the wings of plane and the wheels of car. 
As we increase ${N_r}$ from 2 to 32, the object is slowly completed without huge overlaps between different regions.

\begin{figure}[!t]
  \centering
    \includegraphics[width=\linewidth]{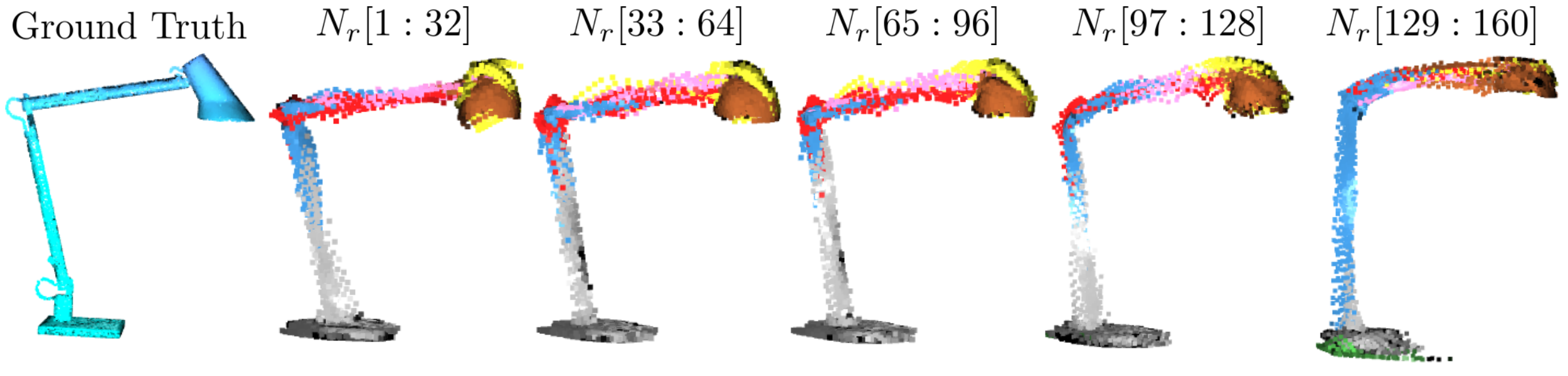}
    \caption{Results when choosing different ranges of rows from $\mathbf{F}'$ to form $\mathbf{F}^*$ instead of selecting the first $N_r =$ 32 rows.}
    \label{fig:less_important}
\end{figure}


In addition to the first 32 rows when setting $N_r$, we also looked into the rows beyond 32. The lamp in \figref{fig:less_important} focuses on the following ranges: $[33:64]$, $[65:96]$, $[97:128]$ and $[129:169]$.
Although the shape of the lamp starts to deform as we go beyond 32, our reconstruction results still captures its overall shape even when we select the range $[129:169]$.
Therefore, this proves that our feature is not constrained to the first 32 rows when sorting and demonstrates the robustness of our softpool feature.


\section{Ablation study on $\tau$}

When computing for $\mathcal{L}_\text{boundary}$, we introduced the threshold $\tau$ to compute the sets. In \tabref{tab:threshold}, we then evaluate different values of $\tau$ and investigate its behavior with respect to the Chamfer distance.
The table demonstrates that the results are not sensitive to the $\tau$, where the thresholds between 0.2-0.9 generate a small difference in the Chamfer distance (with less than 1) from the chosen threshold of 0.3. Notably, compared to the related work, any threshold between 0.1 to 0.9 outperforms the other methods.

\begin{table*}[!h]
\centering
\begin{tabular}{l|ccccccccc}
\toprule	
 $\tau$
 & ~~0.1~~ & ~~0.2~~ & ~~0.3~~ & ~~0.4~~ & ~~0.5~~ & ~~0.6~~ & ~~0.7~~ & ~~0.8~~ & ~~0.9~~ \\
\midrule 
     Chamfer Distance~~ & 7.08 & 5.99 & \textbf{5.94} & 6.12 & 6.19 & 6.18 & 6.21 & 6.25 & 6.71 \\
\bottomrule
\end{tabular}
\caption{Sensitivity of the average Chamfer distance (multiplied by $10^3$) to the threshold $\tau$.
\label{tab:threshold}
}
\end{table*}

\section{Ablation study on $N_r$, $N_r$ and $\mathbf{L}_{\text{boundary}}$}





We investigate the influence of increasing the weight of $\mathcal{L}_\text{boundary}$ on the reconstruction as we change the number of regions $N_f$ and the number selected rows $N_r$. While we chose the best option with $N_r$ set to 8 and $N_r$ set to 32, \tabref{tab:shapenet_regions} also shows that a larger weight on $\mathcal{L}_{\text{boundary}}$ improves the performance when the number of regions is larger, \eg when $N_f$ is 32.

\begin{table*}[!h]
\centering
    \begin{tabular}{l|ccccc}
        \toprule
        $(N_f, N_r)$ & ~~~~(2, 128)~~~~ & ~~~~(4, 64)~~~~ & ~~~~(8, 32)~~~~ & ~~~~(16, 16)~~~~ & ~~~~(32, 8)~~~~ \\
        \midrule
        1 $ \times \mathcal{L}_{\text{boundary}}$ & 7.80 & 6.31 & 5.94 & 6.27 & 6.75 \\
        2 $ \times \mathcal{L}_{\text{boundary}}$ & 7.80 & 6.31 & \textbf{5.91} & 6.25 & 6.72 \\
        10 $ \times \mathcal{L}_{\text{boundary}}$ & 7.82 & 6.29 & 5.95 & 6.01 & 6.19 \\
        \bottomrule
        \end{tabular}
\caption{Influence of $N_f$, $N_r$ and the weight of $\mathcal{L}_{\text{boundary}}$ for object completion on the average Chamfer distance (multiplied by $10^3$).}
\label{tab:shapenet_regions}
\end{table*}

%
%
\bibliographystyle{splncs04}
\bibliography{egbib}